\documentclass[10pt, a4paper]{article}
\usepackage{ltc05}
\usepackage{amsmath}
\usepackage{amssymb}
\usepackage{graphicx}
\usepackage{hyperref}
\usepackage{mathtools}

\title{Fine-tuning Tree-LSTM for phrase-level sentiment classification on a Polish dependency treebank. Submission to PolEval task 2}

\name{Tomasz Korbak$^{{\ast}{\dagger}}$, Paulina \.Zak$^{\ddagger}$}

\address{ $^{\ast}$
Institute of Philosophy and Sociology, Polish Academy of Sciences \\
Nowy \'Swiat 72, 00-330 Warsaw, Poland \\
tkorbak@ifispan.waw.pl \\
\\              
$^{\dagger}$University of Warsaw \\ 
Krakowskie Przedmie\'scie 26/28, 00-927 Warsaw, Poland \\ \\
$^{\ddagger}$Independent researcher \\
paulina.zak1@gmail.com
}

\abstract{We describe a variant of Child-Sum Tree-LSTM deep neural network \cite{tai2015} fine-tuned for working with dependency trees and morphologically rich languages using the example of Polish. Fine-tuning included applying a custom regularization technique (zoneout, described by \cite{krueger2016}, and further adapted for Tree-LSTMs) as well as using pre-trained word embeddings enhanced with sub-word information \cite{bojanowski2016}. The system was implemented in PyTorch and evaluated on phrase-level sentiment labeling task as part of the PolEval competition.}

\begin{document}

\maketitleabstract

\section{Introduction}

In this article, we describe a variant of Tree-LSTM neural network \cite{tai2015} for phrase-level sentiment classification. The contribution of this paper is evaluating various strategies for fine-tuning this model for a morphologically rich language with relatively loose word order -- Polish. We explored the effects of several variants of regularization technique known as zoneout \cite{krueger2016} as well as using pre-trained word embeddings enhanced with sub-word information \cite{bojanowski2016}.

The system was evaluated in PolEval competition. PolEval is a SemEval-inspired evaluation campaign for natural language processing tools for Polish.\footnote{http://poleval.pl} The task that we undertook was phrase-level sentiment classification, i.e. labeling the sentiment of each node in a given dependency tree. The dataset format was analogous to the seminal Stanford Sentiment Treebank\footnote{https://nlp.stanford.edu/sentiment/} for English as described in \cite{socher2013}.

The source code of our system is publicly available under github.com/tomekkorbak/treehopper.

\section{Phrase-level sentiment analysis}

Sentiment analysis is the task of identifying and extracting subjective information (attitude of the speaker or emotion she expresses) in text. In a typical formulation, it boils down to classifying the sentiment of a piece of text, where sentiment is understood as either binary (positive or negative) or multinomial label and where classification may take place on document level or sentence level. This approach, however, is of limited effectiveness in case of texts expressing multiple (possibly contradictory) opinions about multiple entities (or aspects thereof) \cite{thet10}. What is needed is a more fine-grained way of assigning sentiment labels, for instance to phrases that build up a sentence.

Apart from aspect-specificity of sentiment labels, another important consideration is to account for the effect of syntactic and semantic composition on sentiment. Consider the role negation plays in the sentence ``The movie was not terrible": it flips the sentiment label of the whole sentence around \cite{socher2013}. In general, computing the sentiment of a complex phrase requires knowing the sentiment of its subphrases and a procedure of composing them. Applying this approach to full sentences requires a tree representation of a sentence.

PolEval dataset represents sentences as dependency trees. Dependency grammar is a family of linguistics frameworks that model sentences in terms of tokens and (binary, directed) relations between them, with some additional constraint: there must be a single root node with o incoming edges and each non-root node must have a single incoming arc and a unique path to the root node. What this entails is that each phrase will have a single head that governs how its subphrases are to be composed \cite{jurafsky2000}. 

PolEval dataset consisted of a 1200 sentence training set and 350 sentence evaluation test. Each token in a sentence is annotated with its head (the token it depends on), relation type (i.e. coordination, conjunction, etc.) and sentiment label (positive, neural, negative). For an example, consider fig. 1.

\begin{figure}[ht]
\begin{center}
\includegraphics[width=\linewidth]{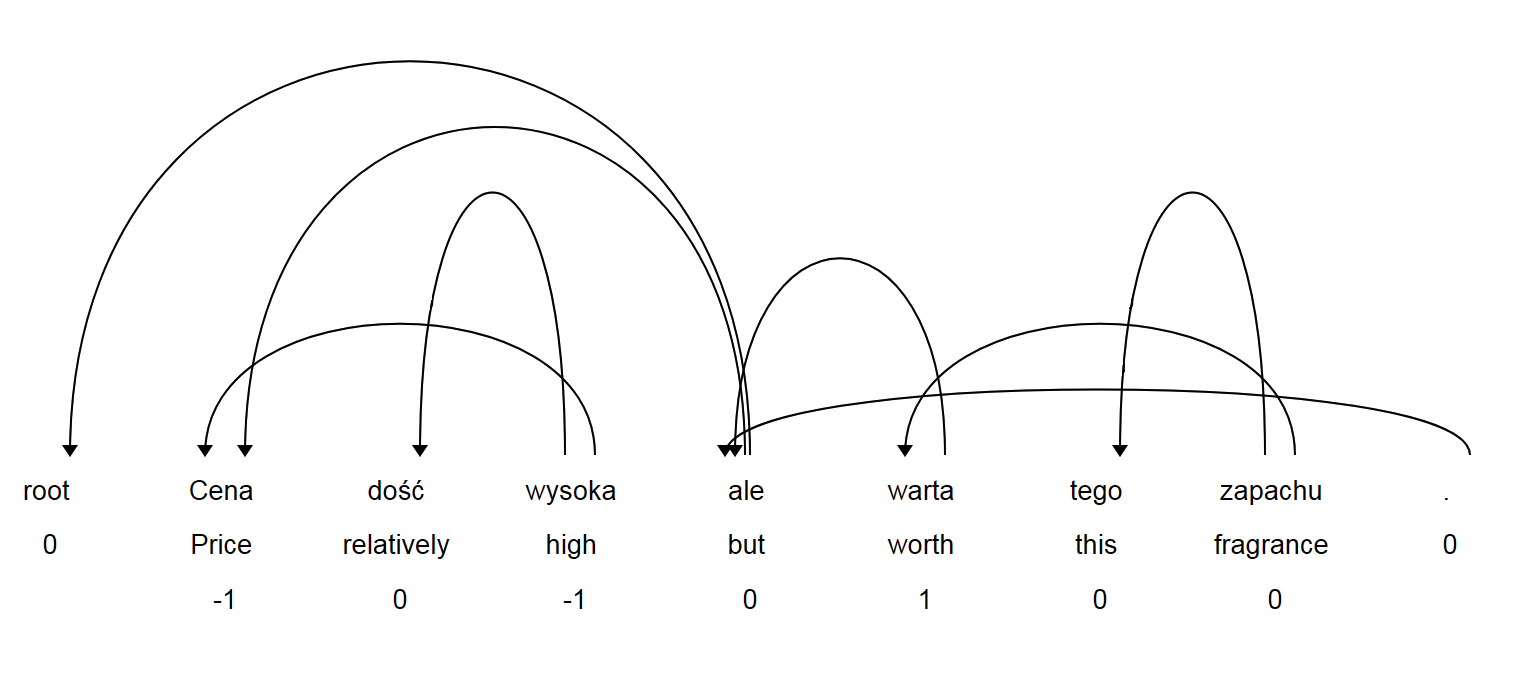}
\caption{An entry in Poleval dataset consists of (1) an ordered list of tokens, (2) dependency relations between them, (3) types of these relations (not used by our model, hence not shown) and (4) sentiment labels for each head (-1, 0, 1).}
\end{center}
\end{figure}

\section{LSTM and Tree-LSTM neural networks}

\subsection{Recurrent neural networks}

A recurrent neural network (RNN) is a machine learning model designed to handle sequential data. It can be described as a dynamical system with transition function $f$:
\begin{gather}\label{rnn-ds}
h_t = f(h_t, x_t; \theta)
\end{gather}
where $h_t$ denotes hidden state at time-step $t$, $x_t$ denotes $t$-th sample and $\theta$ denotes model parameters (weight matrices).

The output $\hat{y}_t$ is then a function of current hidden state $h_t$, current sample $x_t$ and parameters $\theta$:
\begin{gather}\label{rnn-pred}
\hat{y}^{(t)} = g(h^{(t)}, x^{(t)}; \theta)
\end{gather}

In the most simple case (known as Vanilla RNN, or Elman network, cf. \cite{elman1990}), both $f$ and $g$ can be defined as an affine transformations of a concatenation of hidden states and inputs, $[h^{(t)}, x^{(t)}]$, that is:

\begin{gather}
	f(h_t, x_t; \theta) = W_h [h_t, x_t] + b_h
\end{gather}
\begin{gather}\label{rnn-pred-elman}
	g(h_t, x_t; \theta) = W_y [h_t, x_t] + b_y
\end{gather}

for some $W_h, W_y, b_h, b_y \in \theta$. Importantly, none of these parameters depends on $t$; they are shared across time-steps.

\subsection{LSTM cells and learning long-term dependencies}

Thanks to recurrent connections, RNNs are capable of maintaining a working memory (or short-term memory, as opposed to long-term memory captured in weights of forward connections) for storing information about earlier time-steps and use it for classifying subsequent ones. One problem is that the distance between two time-steps has a huge effect on learnability of constraints they impose on each other. This particular problem with long-term dependencies is known as \textit{vanishing gradient problem} \cite{bengio94}.

Long short-term memory (LSTM) architecture \cite{Hochreiter97} was designed address to the problem of vanishing gradient by enforcing constant error flow across time-steps. This is done by introducing a structure called \textit{memory cell}; a memory cell has one self-recurrent connection with constant weight that carries short-term memory information through time-steps. Information stored in memory cell is thus relatively stable despite noise, yet it can be superimposed with each time-step. This is regulated by three gates mediating memory cell with inputs and hidden states: input gate, forget gate and output get.

For time-step $t$, let input gate $i_t$, forget gate $f_t$ and output gate $o_t$ be defined in terms of the following equations (\ref{input-gate}-\ref{output-gate}):

\begin{equation}\label{input-gate}
i_t = \sigma (W^{(i)} x^{(t)} + U^{(i)} h_{t-1})
\end{equation}
\begin{equation}\label{forget-gate}
f_t = \sigma (W^{(f)} x^{(t)} + U^{(f)} h_{t-1})
\end{equation}
\begin{equation}\label{output-gate}
o_t = \sigma (W^{(o)} x^{(t)} + U^{(o)} h_{t-1})
\end{equation}
where $W^{(i)}, W^{(f)}, W^{(o)}$ and $U^{(i)}, U^{(f)}, U^{(o)}$ denote weight matrices for input-to-cell (where input is $x_t$) and hidden-to-cell (where hidden layer is $h_t$) connections, respectively, for input gate, forget gate and output gate. $\sigma$ denotes the sigmoid function.

Gates are then used for updating short-term memory. Let new memory cell candidate $\widetilde{c}_t$ at time-step $t$ be defined as
\begin{equation}\label{new-cell}
\widetilde{c}_t = \tanh (W^{(c)} x_t + U^{(c)} h_{t-1})
\end{equation}
where $W^{(c)}, U^{(c)}$, analogously, are weight matrices for input-to-cell and hidden-to-cell connections and where $\tanh$ denotes hyperbolic tangent function.

Intuitively, $\widetilde{c}_t$ can be thought of as summarizing relevant information about word-token $x_t$. Then, $\widetilde{c}_t$ is used to update $c_t$, according to forget gate and input gate. 
\begin{equation}\label{final-cell}
c_t = f_t \circ c_{t-1} + i_t \circ \widetilde{c}_t
\end{equation}
where $A \circ B$ denotes the Hadamard product of two matrices, i.e. element-wise multiplication.


Finally, $c_t$ is used to compute next hidden state $h_t$, again depending on output gate (defined in equation \ref{output-gate}) that takes into account input and hidden states at current time-step.
\begin{equation}\label{hidden}
h_t = o_t \circ \tanh (c_t)
\end{equation}

In a sequence labeling task, $h_t$ is then used to compute label $\hat{y}_t$ as defined by eq. \ref{rnn-pred-elman}. The forward-propagation for a LSTM network is done by recursively applying equations \ref{input-gate}-\ref{hidden} while incrementing $t$.

\subsection{Recursive neural networks and tree labeling}
\label{tree-labeling}

Recursive neural networks, or tree-structured neural networks, make a superset of recurrent neural networks, as their computational graphs generalize computational graphs of recurrent neural network from a chain to a tree. Whereas a recurrent neural networks hidden state $h_t$ depends only on one previous hidden states, $h_{t-1}$, a hidden state of a recursive neural network depends on a set of descending hidden states $C(h_t)$, when $C(j)$ denotes a set of children of a node $j$.

Tree-structured neural networks have a clear linguistic advantage over chain-structured neural networks: trees make a very natural way of representing the syntax of natural languages, i.e. how more complex phrases are composed of simpler ones.\footnote{Although recursive neural networks are used primarily in natural language processing, they were also applied in other domains, for instance scene parsing \cite{socher2011}.} Specifically, in this paper we will be concerned with a tree labeling task, which is analogous generalization of sequence labeling to tree-structured inputs: each node of a tree is assigned with a label, possibly dependent on all of its children.

\subsection{Tree-LSTMs neural networks}

A Tree-LSTM (as described by \newcite{tai2015} is a natural combination of the approaches described in two previous subsections. Here we will focus on a particular variant of Tree-LSTM known as Child-Sum Tree-LSTM. This variant allows a node to have an unbounded number of children and assumes no order over those children. Thus, Child-Sum Tree-LSTM is particularly well-suited for dependency trees.\footnote{The other variant described by \cite{tai2015}, $N$-ary Tree-LSTM assumes that each node has at most $N$ children and that children are linearly ordered, making it natural for (binary) dependency trees. The choice between these two variant really boils down to the syntactic theory we assume for representing sentences. As PolEval dataset assumes dependency grammar, we decided to go along with Child-Sum Tree-LSTM.}

Let $C(j)$ again denote the set of children of the node $j$. For a given node $j$, Child-Sum Tree-LSTM takes as inputs vector $x_j$ and hidden states $h_k$ for every $k \in C(j)$. The hidden state $h_j$ and cell state $c_j$ are computed using the following equations:

\begin{equation}\label{sum-tlstm}
	\widetilde{h}_j = \sum_{k \in C(j)}^{} h_k
\end{equation}

\begin{equation}\label{input-gate-tlstm}
	i_j = \sigma(W^{(i)}x_j + U^{(i)} \widetilde{h}_j + b_j)	
\end{equation}

\begin{equation}\label{forget-gate-tlstm}
	f_{jk} = \sigma (W^{(f)} x_j + U^{(f)} \widetilde{h}_j + b_f)
\end{equation}

\begin{equation}\label{output-gate-tlstm}
	o_j = \sigma (W^{(o)} x_j + U^{(o)} \widetilde{h}_j + b_o)
\end{equation}

\begin{equation}\label{update-gate-tlstm}
	u_j = \tanh (W^{(u)} x_j + U^{(u)} \widetilde{h}_j + b_u)
\end{equation}

\begin{equation}\label{cell-tlstm}
	c_j = i_j \circ u_j + \sum_{k \in C(j)}^{} f_{jk} \circ c_k
\end{equation}

\begin{equation}\label{hidden-tlstm}
	h_j = o_j \circ \tanh{(c_j)}
\end{equation}

Eqs. \ref{input-gate-tlstm}-\ref{hidden-tlstm} are analogous to eqs. \ref{input-gate}-\ref{final-cell}; they correspond to applying input gate, forget gate, output gate, update gate and computing cell and hidden states. 

In a tree labeling task, we will additionally have an output function

\begin{equation}
	\hat{y}_j = W^{(y)} h_j + b_y
\end{equation}
for computing a label of each node.

\section{Experiments}

We choose to implement our model in PyTorch\footnote{http://pytorch.org/} due to convenience of using a dynamic computation graphs framework.

We evaluated our model on tree labeling as described in subsection \ref{tree-labeling} using PolEval 2017 Task 2 dataset. (For an example entry, see fig. 1).

\subsection{Regularizing with zoneout}\label{zoneout}

Zoneout \cite{krueger2016} regularization technique is a variant of dropout \cite{Srivastava2014} designed specifically for regularizing recurrent connections of LSTMs or GRUs. Dropout is known to be successful in preventing feature co-adaptation (also known as overfitting) by randomly applying a zero mask to the outputs of a given layer. More formally,

\begin{equation}
	h \coloneqq d_t \circ h
\end{equation}
where $d_t$ is a random mask (a tensor with values sampled from Bernoulli distribution). 

However, dropout usually could not be applied to recurrent hidden and cell states of LSTMs, since aggregating zero mask over a sufficient number of time-steps effectively zeros them out. (This is reminiscent of the vanishing gradient problem).

Zoneout addresses this problem by randomly swapping the current value of a hidden state with its value from a previous time-step rather than zeroing it out. Therefore, contrary to dropout, gradient information and state information are more readily propagated through time. Zoneout has yielded significant performance improvements on various NLP tasks when applied to cell and hidden states of LSTMs. This can be understood as substituting eqs. \ref{new-cell}, \ref{hidden} with the following ones:

\begin{equation}
	c_t \coloneqq d^c_t \circ c_t + (1-d^c_t ) \circ c_{t-1}
\end{equation}
\begin{equation}
	h_t \coloneqq d^h_t \circ h_t + (1-d^h_t ) \circ h_{t-1}
\end{equation}
where 1 denotes a unit tensor and $d^c_t$ and $d^h_t$ are random, Bernoulli-sampled masks for a given time-step.

Notably, zoneout was originally designed with sequential LSTMs in mind. We explored several ways of adapting it to tree-structured LSTMs. We will consider only hidden state updates, since cell states updates are isomorphic.

As Tree-LSTM's nodes are no longer linearly ordered, the notion of previous hidden states must be replaced with the notion of hidden states of children nodes. The most obvious approach, that we call ``sum-child" will be randomly replacing the hidden states of node $j$ with the sum of its children nodes' hidden states, i.e.

\begin{equation}
	h_j \coloneqq d^h_j \circ h_j + (1-d^h_j ) \circ \sum_{k \in C(j)}^{} h_k
\end{equation}

Another approach, called ``choose-child" by us, is to randomly choose a single child to replace the node with. 

\begin{equation}
	h_j \coloneqq d^h_j \circ h_j + (1-d^h_j ) \circ h_k
\end{equation}
where $k$ is a random number sampled from indices of the members of $C(j)$.

Apart from that, we explored different values for $d^h$ and $d^c$ as well as keeping a mask fixed across time-steps, i.e. $d_t$ being constant for all $t$.

\subsection{Using pre-trained word embeddings}

Standard deep learning approaches to distributional lexical semantics (e.g. word2vec, \cite{mikolov13}) were not designed with agglutinative languages, like Polish, in mind and cannot take advantage of compositional relation between words. Consider the example of ``chodzi\l{}em" and ``chodzi\l{}am" (Polish masculine and feminine past continuous forms of ``walk", respectively). The model has no sense of morphological similarity between these words and has to infer it from distributional information itself. This poses a problem when the number of occurrences of a specific orthographic word form is small or zero and some Polish words can have up to 30 orthographic forms (thus, the effective number of occurrences is 30 times smaller than the number of occurrences when counting lemmas).

One approach we explore is to use word embeddings pre-trained on lemmatized data. The other, more promising approach, is take advantage of morphological information by enhancing word embeddings with subword information. We evaluate fastText word vectors as described by \cite{bojanowski2016}. Their work extends the model of \cite{mikolov13} with additional representation of morphological structure as a bag of character-level $n$-gram (for $3 \leq n \leq 6$). Each character $n$-gram has its own vectors representations and the resulting word embeddings is a sum of the word vector and its character vectors. Authors have reported significant improvements in language modeling tasks, especially for Slavic languages (8\% for Czech and 13\% for Russian; Polish was not evaluated) compared to pure word2vec baseline.

\section{Results}

We conducted a thorough grid search on a number of other hyperparameters (not reported here in detail due to spatial limitations). We found out that the best results were obtained with minibatch size of 25, Tree-LSTM hidden state and cell state size of 300, learning rate of 0.05, weight decay rate of 0.0001 and L2 regularization rate of 0.0001. No significant difference was found between Adam \cite{kingma2014} and Adagrad \cite{duchi2011} optimization algorithms. It takes between 10 and 20 epochs for the system to converge.

Here we focus on two fine-tunings we introduced: fastText word embeddings and zoneout regularization.

The following word embeddings model were used:
\begin{itemize}
	\item word2vec \cite{mikolov13}, 300 dimensions, pre-trained on Polish Wikipedia and National Corpus of Polish \cite{nkjp08} using lemmatized word forms. Lemmatization was done using Concraft morphosyntactic tagger \cite{waszczuk2012}.
	\item word2vec \cite{mikolov13}, same as above, but using orthographical word forms.
	\item fastText \cite{bojanowski2016}, 300 dimensions, pre-trained on Polish Wikipedia using orthographical word forms and sub-word information.
\end{itemize}

Our results for different parametrization of pre-trained word embeddings and zoneout are shown in tables \ref{emb-table} and \ref{zoneout-table}, respectively. The effects of word embeddings and zoneout were analyzed separately, i.e. results in table \ref{emb-table} were obtained with no zoneout and results in table  \ref{zoneout-table} were obtained with best word embeddings, i.e. fastText.

Note that these results differ from what is reported in official PolEval benchmark. Our results as evaluated by organizing committee, reported in table \ref{originals-table}, left us behind the winner (0.795) by a huge margin. This was due to a bug in our implementation, which was hard to spot as it manifested only in inference mode. The bug broke mapping between word tokens and weights in our embedding matrix. All results reported in tables \ref{emb-table} and \ref{zoneout-table} were obtained after fixing the bug (the model trained on training dataset and evaluated on evaluation dataset, after ground truth labels were disclosed). Note that these results beat the best reported solution by a small margin.

\begin{table}[h]
\begin{center}
\begin{tabular}{lrr}
      \hline
      emb lr & ensemble epochs & accuracy\\\hline\hline
      0.2 & 1 & 0.678 \\\hline
      0.1 & 1 & 0.671 \\\hline
      0.1 & 3 & 0.670 \\\hline
\end{tabular}
\caption{Results of our faulty solution as evaluated by PolEval organizing committee. ``Ensemble epochs" means the number of training epochs we averaged the weights over to obtain a snapshot-based ensemble model.}\label{originals-table}
 \end{center}
\end{table}

\begin{table}[h]
\begin{center}
\begin{tabular}{lrrr}
      \hline
      word embeddings & emb lr & accuracy & time \\\hline\hline
      word2vec, orthographic & 0.0 & 0.7482 & 20:52 \\\hline
      word2vec, orthographic & 0.1 & 0.7562 & 20:26 \\\hline
      word2vec, lemmatized & 0.0 & 0.7536 & 20:01 \\\hline
      word2vec, lemmatized & 0.1 & 0.7737 & 20:09 \\\hline
      fastText, orthographic & 0.0 & 0.8011 & 20:04 \\\hline
      fastText, orthographic & 0.1 & 0.7993 & 20:17 \\
      \hline
\end{tabular}
\caption{A comparison of the effect of pre-trained word embedding on model's accuracy. ``emb lr" means learning rate of the embedding layer, i.e. 0.0 means the layer was kept fixed and not optimized during training. ``time" means wall-clock time of training on a CPU measured in minutes.}
\label{emb-table}
 \end{center}
\end{table}

\begin{table}[h]
\begin{center}
\begin{tabular}{llrrr}
\hline
mask & strategy & $d^c_j$ & $d^h_j$ & accuracy \\
\hline\hline
   n/a &     n/a &  0.00 &  0.00 &  0.8000 \\\hline
   common &     sum-child &  0.01 &  0.00 &  0.8008 \\\hline
   common &     sum-child &  0.00 &  0.01 &  0.8013 \\\hline
 distinct &     sum-child &  0.01 &  0.00 &  0.8013 \\\hline
   common &  choose-child &  0.01 &  0.00 &  0.8015 \\\hline
 distinct &     sum-child &  0.25 &  0.00 &  0.8018 \\\hline
 distinct &  choose-child &  0.01 &  0.01 &  0.8032 \\\hline
 distinct &  choose-child &  0.01 &  0.25 &  0.8051 \\\hline
   common &  choose-child &  0.25 &  0.00 &  0.8052 \\\hline
 distinct &     sum-child &  0.25 &  0.01 &  0.8070 \\\hline
\end{tabular}
\caption{Results extracted from a grid search over zoneout hyperparameters. ``Mask" denotes the moment mask vector is sampled from Bernoulli distribution: ``common" means all node share the same mask, while ``distinct" means mask is sampled per node. ``Strategy" means zoneout strategy as described in section \ref{zoneout}. ``$d^c_j$" and ``$d^h_j$" mean zoneout rates for, respectively, hidden and cell states of a Tree-LSTM. No significant differences in training time were observed.}
\label{zoneout-table}
\end{center}
\end{table}

\section{Conclusions}

As far as word2vec embeddings are concerned, both training on lemmatized word forms and further optimizing embedding yielded small improvements; the two effects being cumulative. FastText vectors, however, beat all word2vec configurations by a significant margin. This result is interesting as fastText embeddings were originally trained on a smaller corpus (Wikipedia, as opposed to Wikipedia+NKJP in the case of word2vec).

When it comes to zoneout, it barely affected accuracy (improvement of about 0.6 percentage point) and we did not found a hyperparameter configuration that stands out. More work is needed to determine whether zoneout could yield robust improvements for Tree-LSTM.

Unfortunately, our system did not manage to win the Task 2 competition, this being due to a simple bug. However, our results obtained after the evaluation indicate that it was very promising in terms of overall design and in fact, could beat other participants by a small margin (if implemented correctly). We intend to prepare and improve it for the next year's competition having learned some important lessons on fine-tuning and regularizing Tree-LSTMs for sentiment analysis.

\section{Acknowledgements}

The work of Tomasz Korbak was supported by Polish Ministry of Science and Higher Education grant DI2015010945 within ``Diamentowy Grant" programme (2016-2020).

\bibliographystyle{ltc05}
\bibliography{references}

\end{document}